\documentclass[sigconf]{acmart}
\renewcommand\footnotetextcopyrightpermission[1]{}
\usepackage{graphicx}
\usepackage{float}
\usepackage{subfigure}
\usepackage{booktabs} 
\usepackage{multirow}
\usepackage{hhline}
\usepackage{makecell}
\usepackage{xcolor}
\usepackage{algorithm}              
\usepackage{algorithmic}
\def\ie{\textit{i.e.}\xspace}

\AtBeginDocument{%
  \providecommand\BibTeX{{%
    \normalfont B\kern-0.5em{\scshape i\kern-0.25em b}\kern-0.8em\TeX}}}

\newcommand{\para}[1]{{\vspace{3pt} \bf \noindent #1 \hspace{0pt}}}
\setcopyright{acmcopyright}
\copyrightyear{2018}
\acmYear{2018}
\acmDOI{10.1145/1122445.1122456}

\begin{document}

\title{Causality Enhanced Origin-Destination Flow Prediction in Data-Scarce Cities}

\author{Tao Feng}
\affiliation{
  \institution{Department of Electronic Engineering, Tsinghua University}
  \city{Beijing}
  \country{China}}

\author{Yunke Zhang}
\affiliation{
  \institution{Department of Electronic Engineering, Tsinghua University}
  \city{Beijing}
  \country{China}}

\author{Huandong Wang}
\affiliation{
  \institution{Department of Electronic Engineering, Tsinghua University}
  \city{Beijing}
  \country{China}}

\author{Yong Li}
\affiliation{
  \institution{Department of Electronic Engineering, Tsinghua University}
  \city{Beijing}
  \country{China}}

\begin{abstract}
Accurate origin-destination (OD) flow prediction is of great importance to developing cities,
as it can contribute to optimize urban structures and layouts.
However, with the common issues of missing regional features and lacking OD flow data,
it is quite daunting to predict OD flow in developing cities.
To address this challenge, we propose a novel Causality-Enhanced OD Flow Prediction (CE-OFP),
a unified framework that aims to transfer urban knowledge between cities and
achieve accuracy improvements in OD flow predictions across data-scarce cities.
In specific, we propose a novel reinforcement learning model to discover universal causalities among urban features in data-rich cities
and build corresponding causal graphs.
Then, we further build Causality-Enhanced Variational Auto-Encoder (CE-VAE) to incorporate causal graphs for effective feature reconstruction in data-scarce cities.
Finally, with the reconstructed features, we devise a knowledge distillation method with a graph attention network to migrate the OD prediction model from data-rich cities to data-scare cities.
Extensive experiments on two pairs of real-world datasets validate that the proposed CE-OFP remarkably outperforms state-of-the-art baselines,
which can reduce the RMSE of OD flow prediction for data-scarce cities by up to 11\%.
\end{abstract}

\maketitle

\section{Introduction} \label{Sec:intro}
\def\ie{\textit{i.e.}\xspace}
Origin-Destination (OD) flow refers to the amount of population flow between two specific regions (\textit{i.e.},
the origin region and the destination region) in a city~\cite{lenormand2016systematic,rossi2007urban}.
As it reflects complex interactions between the urban structure and the travel demand of people, OD flow prediction has been widely recognized as a key enabler for a range of urban transportation applications, including transport facilities planning, traffic control and taxi dispatching~\cite{castiglione2015activity,deng2016latent}.

For developing cities, OD flow prediction plays an important role in optimizing the structure and layout of urban regions.
Nevertheless, due the to common issues of lacking OD flow data and missing regional features,
data scarcity has always been a major stumbling block in predicting accurate OD flow in developing cities.
In specific, with the limited number of road sensors and traffic cameras,
there is a high chance of missing features and even flow data in particular regions.
Numerous existing works~\cite{robinson2018machine,pourebrahim2019trip,liu2020learning} have investigated the OD flow prediction problem, but they failed to address the data scarcity problem with careful considerations.
Generally, conventional methods have constrained capabilities in urban OD flow prediction,
as most of them directly follow typical physical laws rather than combining flow data with features of different regions.
With the rapid development of Machine Learning (ML) techniques,
emerging ML models for OD flows exhibit stronger predicting capabilities,
such as decision trees~\cite{robinson2018machine,pourebrahim2019trip} and graph neural networks~\cite{velivckovic2017graph,liu2020learning}.
Nonetheless, these approaches require a large amount of OD flow data to fit the large number of model parameters,
which prevents their applications in data-scarce cities.
The rising paradigm of transfer learning provides us with a promising solution to this problem~\cite{Pan2010Survey}.
Based on the transfer learning techniques,
we can utilize the massive OD flow data available at data-rich cities to help us predict the OD flow of data-scarce cities.
Under the circumstances, the missing regional features at the data-scarce cities become the bottleneck of the OD flow prediction.

In order to solve this problem, in this paper, we focus on predicting the OD flow in data-scarce cities by first reconstructing missing urban regional features based on the observed urban regional features and then implementing OD flow prediction based on them. Specifically, both the feature reconstruction model and the OD flow prediction model are learned by utilizing the massive data in data-rich cities. However, there are several main challenges in developing such models. First, reconstructing missing features requires modeling the relationships between observed features and missing features. However, such relationships obtained from the data-rich cities are hard to generalize to the data-scarce cities due to the differences in cities. Therefore, universal  relationships shared by both of data-rich cities and data-scarce cities  between observed features and missing features  are  needed to address this problem.
Second, due to the differences in the distribution and scale of urban regional features between data-rich cities and data-scarce cities, the feature reconstruction model obtained in data-rich cities should not only  consider the accuracy  but also its uncertainty of the reconstructed features. Traditional reconstruction methods like auto-encoder (AE) ~\cite{semwal2017robust,ng2011sparse} model can reconstruct the features with high accuracy, but they  fail to considering  the uncertainty of the features caused by the differences in the distributions and scales, thus  weakening its ability to generalize to different scenarios.


To solve these challenges, we propose a novel a novel \textbf{Causality-Enhanced OD Flow Prediction (CE-OFP)} model  based on a \textbf{Causality Enhanced Variational Auto-Encoder (CE-VAE)}  to transfer urban knowledge of missing  features and OD flow   from data-rich cities to the data-scarce cities. For the first challenge, we propose to search the causal graph between urban features in data-rich cities as universal relationships that can be shared across multiple cities, which is recognized by many existing researches~\cite{wang2021ordering,zhu2019causal}. Specifically, we  introduces a state-of-art reinforcement learning (RL) based method to model the causal graph searching as a sequential decision making problem, which alleviates the difficulty of NP-hard search process. 
The obtained causal graph depicts the essential universal relationships between urban regional features, which serves as basis for the generalization of feature reconstruction models in different cities.
As for the second challenge, we propose a  Causality Enhanced Variational Auto-Encoder (CE-VAE)  to model the accuracy and uncertainty of missing features. CE-VAE first exploits the universal causal graph of urban regional features to find the correlation paths between observed and missing features in data-scarce cities, which guarantees the generalization ability of the feature reconstruction model across cities. It further  incorporates the causal graph into
the inference process of Variational Auto-Encoder (VAE) and models the accuracy and uncertainty of the missing features by outputting the mean value  and  the variance of the missing features, respectively. 
Based on the the mean value  and  the variance of the missing features,
we  utilize a GAT-based (graph attention neural network) knowledge distillation method to migrate the OD prediction model of data-rich cities to data-scarce cities, thereby enhancing the prediction performance of data-scarce cities.

The contributions of our work can be summarised as follows:
\begin{itemize}

\item  We propose  a novel Causality-Enhanced OD Flow Prediction (CE-OFP) model based on a Causality Enhanced Variational Auto-Encoder (CE-VAE) to transfer urban knowledge of missing features and OD flow from data-rich
cities to the data-scarce cities, which enhances the prediction performance of data-scarce cities.

\item  We propose a Causality Enhanced Variational Auto-Encoder (CE-VAE) feature reconstruction model by organically incorporating urban causal
graph into the inference process of Variational Auto-Encoder to construct a latent representation vector considering the
accuracy and uncertainty of the unobserved features, which promotes its generalization ability in different cities.

\item  Extensive experiments show that our framework performs better than seven state-of-the-art OD flow prediction methods by 11\% in  data-scarce cities.
\end{itemize}
\section{Preliminaries} \label{Sec:preliminaries}
In this section, we will introduce the definition of necessary notations and give the problem formulation of origin-destination flow prediction in data-scarce cities with missing urban regional features.

\subsection{Notation Definition} \label{Sec:notation}

\textbf{Definition 1 (Urban Region)} Urban regions are a series of non-overlapping areas in the city.  Following the previous work~\cite{carey1871principles}, city is divided into irregular urban regions by road network composed of multi-level roads.  We denote the region set as $\mathcal{R}$ and a single region as $r \in \mathcal{R}$. We further use $\mathcal{R}^{src}$ to denote the region set of the data-rich  city and $\mathcal{R}^{tar}$ to denote the region set of the data-scarce city.

\textbf{Definition 2 (Urban Regional Features)} Each region $r$ has its own urban regional features, such as population size, economic development status, etc. We indicate the full set of features using $\mathcal{F} = \{ F_i | i = 1,2,...,|\mathcal{F}| \}$. Due
to the limited number of sensors or cameras in developing cities, urban regional features have a high probability of missing. Therefore, we  further define the set of features that can be observed in the  city  as $X \subseteq \mathcal{F}$ and the set of the missing features  as $Y \subseteq \mathcal{F}$.

\textbf{Definition 3 (Urban Topology)} Urban topology is defined as  adjacency distance matrix of  urban region pairs, which is denoted as $\mathcal{A} = \{ distance(r_i, r_j) | r_i, r_j \in \mathcal{R} \}$. 

\textbf{Definition 4 (OD Flow)} We define the OD flow as the commute number of people who move from one urban region to another~\cite{carey1871principles,stouffer1940intervening}. It is represented as  $\mathcal{M} = \{ M^{o,d}|o \in {\mathcal{R}} ~ and ~ d \in \mathcal{R} \}$. The corresponding city of the set is used as the superscripts to indicate, for example, $\mathcal{M}^{src}$.

\textbf{Definition 5 (Causal Directed Acyclic Graph)} We denote causal Directed Acyclic Graph (DAG) ~\cite{zhu2019causal} as $\mathcal{G}$, whose  nodes  depict the urban regional features and  edges between two features depict their causal relationships.


\subsection{Problem Formulation}
After giving the above key notations, we can propose the mathematical definition of the problem solved in this work as follows:

\textbf{Definition 5 (OD Flow Prediction in Data-scarce Cities with Missing Urban Regional Features)} Given the complete urban regional features of all regions 
$\{ F_i^r | i=1,2,...,|\mathcal{F}| ~ and ~ r \in \mathcal{R}^{src} \}$ and complete origin-destination flow  $\mathcal{M}^{src}$ in the data-rich city  and  observed urban regional features of all region in data-scarce  cities $\{ X_i^r | i=1,2,...,|\mathcal{X}| ~ and ~ r \in \mathcal{R}^{tar} \}$, combined with the urban topology of all cities $\mathcal{A}^{src}$ and $\mathcal{A}^{tar}$, the problem is trying to predict the complete origin-destination flow of the data-scarce  cities $\mathcal{M}^{tar}$.

\section{Methods} \label{Sec:method}
\begin{figure}[h]
  \centering
  \includegraphics[width=\linewidth]{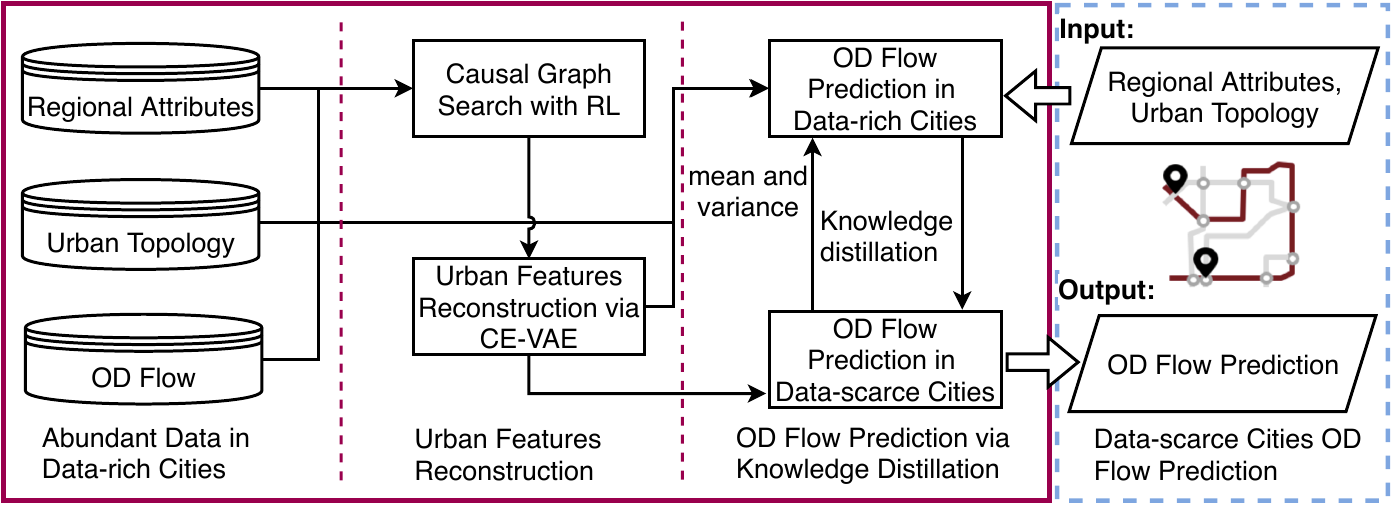}
  \caption{Overview of CE-OFP.}
  \label{Fig:framework}
  \vspace{-2mm}
\end{figure}

An overview of CE-OFP’s architecture is presented in Fig. \ref{Fig:framework}. We first search the causal graph of urban regional features as urban causal knowledge  through a RL-based causal discovery method in data-rich cities. Then we propose the CE-VAE model by incorporating urban causal graph into the inference process of VAE to reconstruct the missing urban features. The mean and variance vector of missing features output by CE-VAE are further used for the OD flow prediction via
knowledge distillation.

\subsection{Causal Graph Search among Urban Features with Reinforcement Learning} \label{sec:causalrl}
Searching for the causal graph among multiple urban regional features  is an NP-hard combinatorial optimization problem~\cite{wang2021ordering,zhu2019causal}, which is difficult to solve by traditional causal discovery methods.  Reinforcement learning (RL) is very powerful for solving such large-scale combinatorial optimization problems~\cite{mazyavkina2021reinforcement,barrett2020exploratory,bello2016neural}.
Therefore, we introduce a state-of-art RL-based  causal discovery method~\cite{wang2021ordering} to search the causal graph among urban regional features.  We first formulate the problem as a one-step Markovian Decision-making Process (MDP). Formally, each MDP can be describe as a  4-tuple  $\left ( \mathit{S},\mathit{A},\mathit{P},\mathit{R} \right )$. Specially, $\mathit{S}$ and $\mathit{A}$ represent the state space and action space respectively. $\mathit{P}:\mathit{S}\times \mathit{A}\rightarrow \mathbb{R}$ represents the probability of state transition. That is, $\mathit{P(s_{t+1}|s_{t},a_{t})}$ is the probability distribution of the next state $s_{t+1}$ conditioned on the current state $s_{t}$ and action $a_{t}$. Finally, $\mathit{R}:\mathit{S}\times \mathit{A}\rightarrow \mathbb{R}$ is the reward function with $\mathit{R(s,a)}$ representing the  reward received by executing the action $a \in \mathit{A}$ under the state $s \in \mathit{S}$. In addition, for the convenience of modeling, we indicate the full set of factors using $\mathcal{F} = \{ F_i | i = 1,2,...,|\mathcal{F}| \}$. Below, we 
detail how to model the above components of the MDP.

\begin{itemize}
    \item \textbf{State:} 
    As mentioned in previous studies~\cite{wang2021ordering,zhu2019causal}, it is difficult to capture the underlying causal relationships through directly using the urban regional features $F$ as the state. These studies also inspire us to use an encoder to embed each factor $F_{i}$ to state $s_{i}$, which is beneficial for the causal discovery   process. Therefore, the state space can be obtained as  $\mathit{S}=(s_{1},s_{2}...,s_{|\mathcal{F}|})=encoder(F_{1},F_{2}...,F_{|\mathcal{F}|})$.
    Motivated by ~\cite{wang2021ordering}, we exploit self-attention based encoder in the Transformer structure  in our reinforcement learning model.

    \item \textbf{Action:}  
     The action of our RL framework is to generate a binary adjacency matrix $U$, which  corresponds to causal DAG $\mathcal{G}$. For example, if the $i$-th row and $j$-th column of $U$ is 1, it means that the $i$-th urban regional feature is the cause for the $j$-th  urban regional feature in the causal DAG $\mathcal{G}$.
     
     \item \textbf{State transition:} 
    In the one-step MDP, the state $s$ will be directly transferred to the end state of the episode after the first action is executed.

    \item \textbf{Reward:} 
     Our optimization goal is to search over the space of all DAGs to find  a $\mathcal{G}$ with the minimum Bayesian Information Criterion score $S_{BIC}(\mathcal{G})$ \cite{watanabe2013widely}, which depicts how well the obtained DAG matches the observed data causally. Here we adopt the DAG constraint $\rho(\mathcal{G})$ proposed by this paper~\cite{wang2021ordering} as a penalty term to add to our optimization goal  to make sure that the $\mathcal{G}$ we search for is a DAG. Therefore, we set the episode reward as $R=-S_{BIC}(\mathcal{G})-\rho(\mathcal{G})$, which will be obtained when executing the binary adjacency matrix $U$. The process of RL learning will maximize reward $R$, thereby minimizing $S_{BIC}(\mathcal{G})$ and  $\rho(\mathcal{G})$ simultaneously.
\end{itemize}

Based on the above MDP models, the  causal discovery is described by a policy function $\pi :\mathit{S} \rightarrow \mathit{A}$. Specially, $\pi(a|s)$ represents the probability of choosing action $a$ under current state $s$.  We adopt a self-attention encoder and an LSTM based decoder  ~\cite{zhu2019causal} to map the state to action. Base on the RL framework, we introduce actor-critic algorithm~\cite{mazyavkina2021reinforcement} to train the our RL framework so as to obtain the best causal graph $\mathcal{G}$, whose  nodes  depict the urban  features and  edges between two factors depict their causal relationships.

\subsection{Urban Features Reconstruction  via Causality Enhanced Variational Auto-encoder}

Urban features reconstruction model is first trained in data-rich cities and then transferred to data-scarce cities, which helps reconstruct the missing features in data-scarce cities. Traditional reconstruction methods like VAE \cite{kingma2013auto} usually 
reconstruct missing features based on the fully connected neural network, which assume that the missing features are all related to the observed features. However, these methods are prone to overfitting in data-rich cities and degrades generalization performance applied in data-scarce cities due to modeling redundant relationships between observed  features and missing features~\cite{krizhevsky2017imagenet,lopez2017conditional,kingma2013auto,sohn2015learning}. To solve this problem, we propose CE-VAE  to incorporate causal graph $\mathcal{G}$  in Section \ref{sec:causalrl} into  feature reconstruction of VAE in both encoder and decoder as shown in Fig. \ref{Fig:encoder} and Fig. \ref{Fig:decoder}, which models essential universal relationships between observed  features and missing features~\cite{wang2021ordering,zhu2019causal}. We will introduce the design of our encoder and decoder in CE-VAE in the following in detail.



\subsubsection{Encoder}

\begin{figure}[t]
  \centering
  \includegraphics[width=\linewidth]{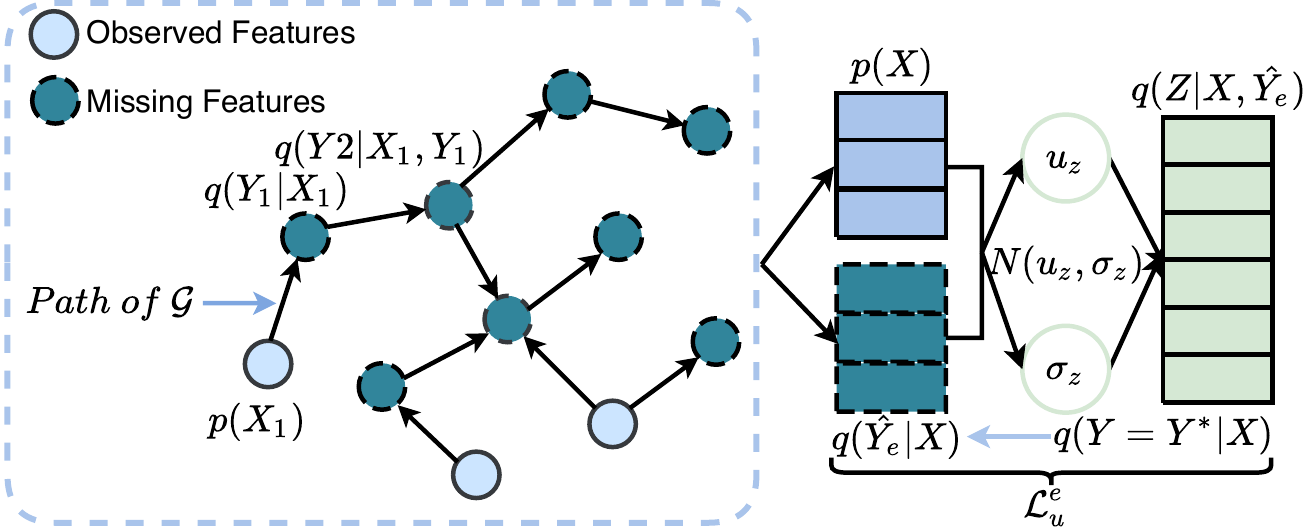}
  \caption{Encoder $q(Z,\hat{Y_e}|X)$ of CE-VAE.}
  \label{Fig:encoder}
  \vspace{-2mm}
\end{figure}

The encoder of our CE-VAE is to learn a conditional probability distribution $q(Z|X)$ that models the dependencies of the hidden embedding $Z$ containing the intact  information of all features on observed urban features $X$ as shown in Fig. \ref{Fig:encoder}. Therefore, we need to reconstruct the missing features $\hat{Y_e}$ basing on  observed urban features $X$ first, and then obtain hidden embedding $Z$ containing complete feature information through $X$ and $\hat{Y_e}$. To reconstruct $\hat{Y_e}$ based on $X$, CE-VAE takes the paths in the causal graph $\mathcal{G}$  as 
reconstruction paths  and reconstructs missing features basing on their parent feature nodes according to the causal graph one by one. For example, as shown in Fig. \ref{Fig:encoder}, CE-VAE first reconstructs the  conditional probability distribution $q(Y_1|X_1)$ of $Y_1 \in \hat{Y_e}$ basing on its parent feature node $X_1$ and then reconstructs the  conditional probability distribution $q(Y_2|X_1,Y_1)$ of $Y_2 \in \hat{Y_e}$ basing on its parent feature nodes $X_1,Y_1$, which is 
according to paths of causal graph $\mathcal{G}$. Following VAE, we further model each feature in encoder via Gaussian probability distribution with its mean and variance. Specifically, each $Y_i \in \hat{Y_e}$ can be described as, 

\begin{equation}
    q(Y_i|PA_{Y_i}^{\mathcal{G}}) \sim \mathcal{N}(\mu_{Y_i}, \sigma_{Y_i}),
\end{equation}

\begin{equation}
    \mu_{Y_i}, \sigma_{Y_i} = MLP({||^{j}_{j\in \{PA_{Y_i}^{\mathcal{G}}\}} {(X_j,Y_j)}}),
\end{equation}
where $||$ denotes the concatenation of features, $\mu_{Y_i}, \sigma_{Y_i}$  denote mean and variance of Gaussian probability distribution $\mathcal{N}$ respectively, $ PA_{Y_i}^{\mathcal{G}}$ denotes the parent feature nodes of node $i$ in $\mathcal{G}$ and MLP denotes neural networks. 

Basing on the observed features $X$ and reconstructed missing features $\hat{Y_e}$, the hidden embedding $Z$ containing the intact  information of all features can be described as, 
 
\begin{equation}
     q(Z|X,\hat{Y_e}) \sim \mathcal{N}(\mu_{Z}, \sigma_{Z}),
\end{equation}

\begin{equation}
    \mu_{Z}, \sigma_{Z} = MLP(X||\hat{Y_e}),
\end{equation}
where $\mu_{Z}, \sigma_{Z}$ denote mean and variance of Gaussian probability distribution $\mathcal{N}$ and model the accuracy and uncertainty of
the missing features respectively.

\subsubsection{Decoder}

\begin{figure}[t]
  \centering
  \includegraphics[width=0.8\linewidth]{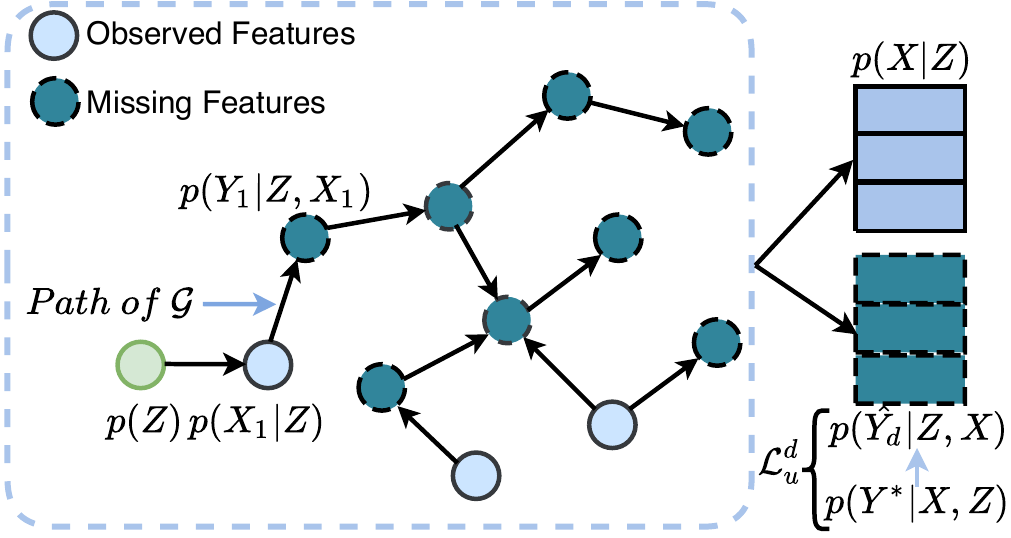}
  \caption{Decoder $p(X,\hat{Y_d}|Z)$ of CE-VAE.}
  \label{Fig:decoder}
  \vspace{-2mm}
\end{figure}

The decoder of CE-VAE is to learn  conditional
probability distributions $p(X|Z)$ and  $p(\hat{Y_d}|Z,X)$  that model the dependencies of
the observed features $X$ and the missing features $\hat{Y_d}$ respectively as shown in Fig. \ref{Fig:decoder}. The design of decoder is similar to encoder, which reconstructs the features one by one according to the paths of causal graph $\mathcal{G}$. For example, as shown in Fig. \ref{Fig:decoder}, CE-VAE first reconstructs the  conditional probability distribution $p(X_1|Z)$ of $X_1$ basing on the hidden embedding  $Z$ and then reconstructs the  conditional probability distribution $p(Y_1|Z,X_1)$ of $Y_1 \in \hat{Y_d}$ basing on the hidden embedding  $Z$   and its parent feature node $X_1$, which is 
according to paths of causal graph $\mathcal{G}$. Following VAE, we set $p(Z)=\mathcal{N}(0, 1)$ as normal Gaussian probability distribution and  further model each feature in decoder via Gaussian probability distribution with its mean and variance. Specifically, each $X_i$ or  $Y_i \in \hat{Y_e}$ (the following formula uses $Y_i$ as an example)  can be described as,

\begin{equation}
    p(\hat{Y_i}|Z,PA_{Y_i}^{\mathcal{G}}) \sim \mathcal{N}(\mu_{Y_i}, \sigma_{Y_i}),
\end{equation}

\begin{equation}
    \mu_{Y_i}, \sigma_{Y_i} = MLP({Z,||^{j}_{j\in \{PA_{Y_i}\}} {(X_j,Y_j)}}),
\end{equation}

\subsubsection{Training}
 Similar to  VAE, the integral optimization objective of our proposed Causal Enhanced Variational Auto-encoder is the Evidence Lower Bound (ELBO) formulated as follows:

\begin{align}
    \mathcal{L}_{ELBO}
    &=\log p(X)- KL[q(Z,Y|X)|p(Z,Y|X)] \\
    &=\log p(X)- \iint q(Z,Y|X)\log \frac{q(Z,Y|X)}{p(Z,Y|X)}dYdZ \\
    &=\log p(X)-\iint q(Z,Y|X)\log \frac{p(X)q(Z,Y|X)}{p(X,Y,Z)}dYdZ \\
    &=-\iint q(Y|X)q(Z|X,Y)\log \frac{q(Z,Y|X)}{p(Z)p(X,Y|Z)}dYdZ \\
    &= E_{q(Z,Y|X)}[\log p(Z)+\log p(X,Y|Z)-\log q(Y|X) \\
    &-\log q(Z|X,Y)] \nonumber,
\end{align}
where $q$ and $p$ denote  conditional probability distributions of encoder and decoder, respectively.

In order to reduce the error accumulation of these features in the reconstruction process, similar to \cite{louizos2017causal}, we will add two extra terms $\mathcal{L}_{u}^e$ and $\mathcal{L}_{u}^d$  in ELBO to endow  $Y$ with physical constraints as shown in Fig. \ref{Fig:encoder} and \ref{Fig:decoder}:
\begin{equation}
    \mathcal{L}_{u} =  \log \textit{q}(Y=Y^*|X) + \log \textit{p}(Y=Y^*|X,Z),
\end{equation}
where the $Y^*$ means the observed labels of the missing features in data-rich cities. This ensures that the reconstruction of $Y$ learned stably to avoid the accumulation of errors and noise when computing sequentially on the causal path. Based on the above formulation,  the final loss function used to optimize the CE-VAE is shown below:
\begin{equation}
    \mathcal{O} = -(\mathcal{L}_{ELBO} + \beta \mathcal{L}_{u}),
\end{equation}
where $\beta$ is the weight of auxiliary loss.


\subsection{Origin-destination Flow Prediction via Knowledge Distillation}

\begin{figure}[h]
  \centering
  \includegraphics[width=\linewidth]{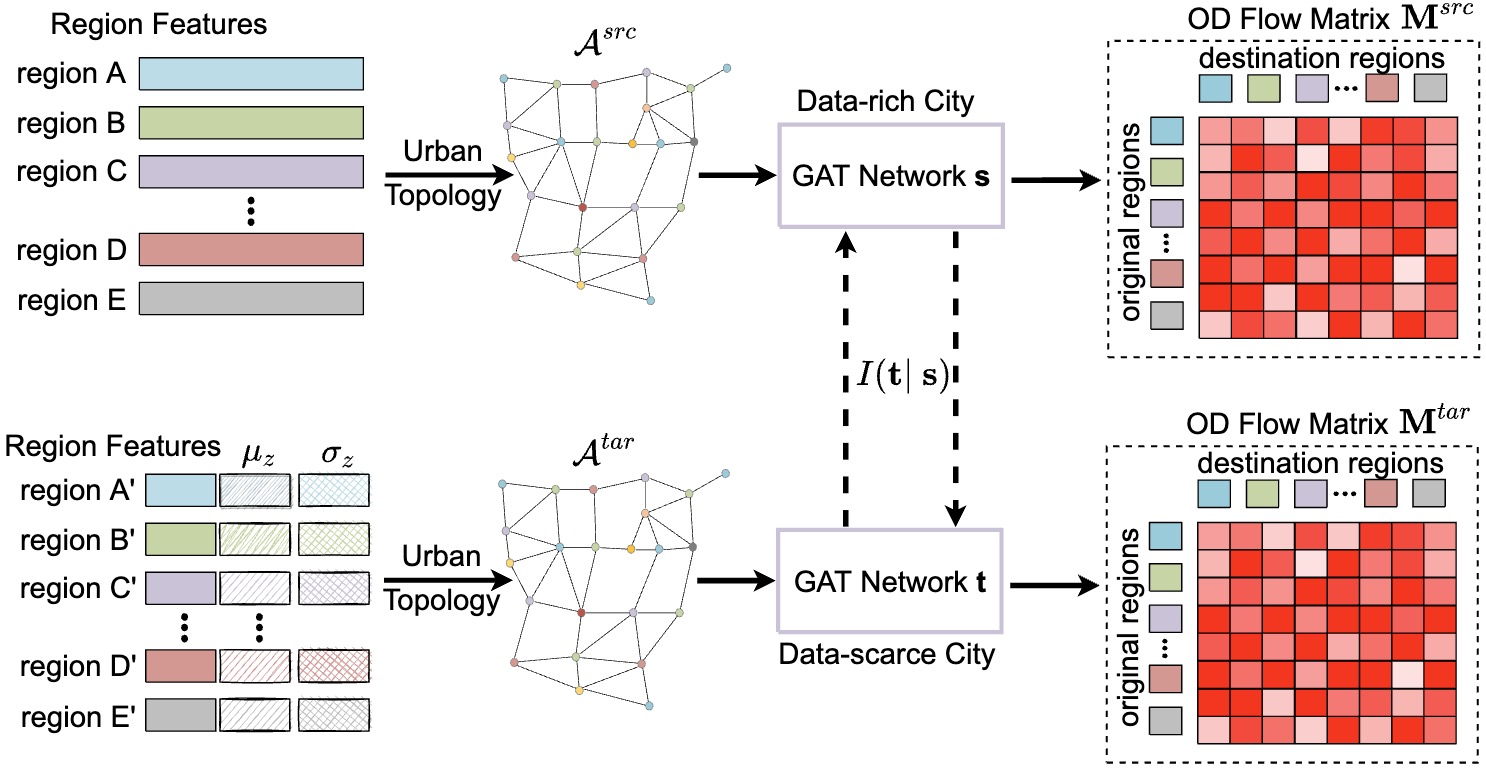}
  \caption{GAT based model for OD flow prediction via
knowledge distillation.}
  \label{Fig:knowledge distillation}
  \vspace{-2mm}
\end{figure}

In this section, we first utilize the learned  $\mu_{z}$ and $\sigma_{z}$ output by the encoder of CE-VAE model in Fig. \ref{Fig:encoder} to make up for the missing urban regional features in data-scarce city. And then we propose a GAT-based (graph attention neural network) model for OD flow prediction via knowledge distillation~\cite{zhang2018deep}. Our model is shown in Fig. \ref{Fig:knowledge distillation}.


We construct a graph network on urban space, with regions as nodes and distances  between regions in urban topology as edges. This is because based on spatial continuity, neighboring regions have some similarities, and using graph neural networks can make full use of the similarity between regions to make the labels propagate over the urban space. We use GAT to extract spatial features of regions on a graph network in urban space and use original features and  features of destination to predict 
OD flow, as shown in Fig. \ref{Fig:knowledge distillation}. 

We use MSE as the loss of gradient descent to predict OD flow, which is calculated by the following Formula.
\begin{equation}
    \setlength{\abovedisplayskip}{0.5pt}
    \setlength{\belowdisplayskip}{-0.5pt}
    {\mathcal{L}_\textit{MSE}}{\;}={\;}{{{\frac{1}{|\textbf{M}|}}{\sum_{i,j}}{||}{{{\textbf{M}}^{r_i,r_j}}-{\hat{\textbf{M}}^{r_i,r_j}}}{{||}_2^2}}},
\end{equation}

Because of the problem of sparse OD flow data in developing cities, we exploit the method of knowledge distillation when transferring the model of prediction to remedy the scarce flow data  in the data-scarce cities. Specifically, we add a loss so that the prediction model of the data-scarce city contains as much information as possible from the model of  data-rich city. The transfer loss is calculated by the following formulas,
\begin{equation}
    \mathcal{L}_{transfer} = I(\textbf{t}|\textbf{s}),
\end{equation}
\begin{equation}
    I(\textbf{t}|\textbf{s}) = \sum_{d=1}^{D}log\sigma_d + \frac{\textbf{t}_d-(\mu_d (\textbf{s}))^2} {2\sigma_{d}^{2}},
\end{equation}
where $I(\textbf{t}|\textbf{s})$ means the discrepancy between the layers output of data-rich city and data-scarce city's model, the $d$ is the output of layers of neural networks and $\mu_d$ and $\sigma_d$ is the corresponding mean and deviation.

Therefore, the final loss of training model in data-scarce city is the addition of MSE and transfer loss.
\begin{equation}
    \mathcal{L}_{pred} = L_{MSE} + L_{transfer}
\end{equation}

We summarize the training process of the CE-OFP algorithm in the Table \ref{tab:all} of Appendix A. We first train CE-VAE using the data of data-rich city as lines (8-14). And then we combine the $\mu_z$ and $\sigma_z$ output by the encoder of the trained CE-VAE with the original regional features to construct
the data in data-scarce city as line (15). Finally, we train our  GAT-based OD flow prediction model via
knowledge distillation to obtain the OD flow in data-scarce city as lines (16-21).

\section{Experiments} \label{Sec:exp}

\begin{table*}[t]
    \centering
    \caption{Overall performance comparison in two scenarios.}
    \resizebox{0.6\textwidth}{!}
  {%
    \begin{tabular}{c||ccc|ccc}
        \Xhline{1pt}
        \multirow{3}{*}{Model}   & \multicolumn{3}{c|}{NYC-Chi}   & \multicolumn{3}{c}{NYC-Sea}  \\
         & RMSE $\downarrow$ & SMAPE $\downarrow$ & CPC $\uparrow$  & RMSE $\downarrow$ & SMAPE$\downarrow$ & CPC $\uparrow$ \\ 
         \hline \hline
        {Gravity Model} & 18.06      & 0.92       & 0.40 & 22.37      & 0.96        & 0.37  \\
        {GBRT}& 9.64      & 0.81       & 0.58   & 17.26      & 0.86        & 0.55 \\
        {GAT}& 10.89       & 0.83       & 0.53 & 16.03      & 0.78        & 0.55\\ 
        {GMEL}& 10.31       & 0.81       & 0.57    & 16.15      & 0.75        & 0.56\\
        \hline \hline
        {MF-GAT}& 10.72 & 0.83 & 0.54  & 16.22 & 0.78 & 0.56 \\
        {AE-GAT}& 9.35 & 0.73 & 0.59  & 16.24 & 0.78 & 0.57 \\
        {VAE-GAT}& 9.39 & 0.73 & 0.58  & 14.97 & 0.77 & 0.57 \\
        \hline \hline
        CE-OFP & \textbf{8.86}  & \textbf{0.69} & \textbf{0.62} & \textbf{14.84} & \textbf{0.59} &\textbf{0.59}\\
         \Xhline{1pt} 
    \end{tabular}}
    \label{tab:overall}
\end{table*}


\subsection{Experimental Settings}

\subsubsection{Datasets}
Following previous studies on OD flow prediction~\cite{robinson2018machine,pourebrahim2019trip,liu2020learning},  we use datasets  collected from  New York City, Chicago and Seattle for evaluation, which include the urban regional features, urban topology and OD flow. For each evaluation scenario, we assume New York City as the data-rich city and Chicago and Seattle as the data-scarce cities. Specifically, we make Chicago and Seattle into data-scarce cities by randomly masking some types of urban regional features and OD flow data following previous work~\cite{Pan2010Survey}. Moreover, we summarize our experimental 
scenarios as \textbf{NYC-Chi} and \textbf{NYC-Sea}. The details of  datasets we use in our experiments are shown as follows:

\begin{itemize}
    \item \textbf{Urban Regional Features}. Following the previous work~\cite{carey1871principles}, we divide each city  into irregular urban regions by road network composed of multi-level census tracts.  The regional features of each region includes two parts: \textit{demographics} and \textit{POI (Points of Interest) distribution}. The demographics data is collected from the United States Census Bureau\footnote{https://www.census.gov.} and POI distribution data is crawled from the OpenStreetMap (OSM) \cite{OpenStreetMap} which is an open source crowdsourced map data collection service.
    \item \textbf{Urban Topology}. We build topological relationship among regions and calculate the distance matrix between each pair of regions based on the administrative map provided by the United States Census Bureau.
    \item \textbf{OD Flow}. The commute OD flow is from the Origin-Destination Employment Statistics organized by the Longitudinal Employer-Household Dynamics (LEHD) program\footnote{https://lehd.ces.census.gov} of the United States Census Bureau.
\end{itemize}

All datasets mentioned above have been compiled by us and will be published with this paper.


\subsubsection{Metrics}
We use Root Mean Square Error (RMSE), Systemitic Mean Absolute Percentage Error (SMAPE), and Common Part of Commuters (CPC) as the metrics of the performance on OD prediction task in our experiments. For feature reconstruction task, we use the log likelihood of reconstructed features (denoted by $L$) and Mean Squared Error (MSE) as the metrics to measure the gap between the reconstructed features and the real features.

\begin{equation}
    \setlength{\abovedisplayskip}{0.5pt}
    \setlength{\belowdisplayskip}{-0.5pt}
    {\textit{RMSE}}{\;}={\;}{\sqrt{{\frac{1}{|\textbf{M}|}}{\sum_{i,j}}{||}{{{\textbf{M}}^{r_i,r_j}}-{\hat{\textbf{M}}^{r_i,r_j}}}{{||}_2^2}}},
\end{equation}
\begin{equation}
    \setlength{\abovedisplayskip}{0.5pt}
    \setlength{\belowdisplayskip}{-0.5pt}
    {\textit{SMAPE}}{\;}={\;}{\frac{100\%}{|\textbf{M}|} \sum_{i,j\in \mathcal{R}^{tar}} \frac{{\textbf{M}^{r_i,r_j}} - {\hat{\textbf{M}^{r_i,r_j}}}}{(|\textbf{M}^{r_i,r_j}|+|\hat{\textbf{M}^{r_i,r_j}}|)/2}},
\end{equation}
\begin{equation}
    \setlength{\abovedisplayskip}{0.5pt}
    \setlength{\belowdisplayskip}{-0.5pt}
    {\textit{CPC}}{\;}={\;} {\frac{2 \sum_{i,j\in \mathcal{R}^{tar}} min(\textbf{M}^{r_i,r_j}, \hat{\textbf{M}^{r_i,r_j}}) } {\sum_{i,j\in \mathcal{R}^{tar}} {\hat{\textbf{M}^{r_i,r_j}}}+ \sum_{i,j\in \mathcal{R}^{tar}} {\textbf{M}^{r_i,r_j}} }},
\end{equation}
where RMSE is commonly used in regression problem, SMAPE shows the prediction error as a percentage of the ground truth and CPC, which measures the common part of the prediction flow and true values, is widely used in research of commuting flow. In addition, log likelihood and MSE measure the distribution difference  and the value difference between our reconstructed features and the real features in our feature reconstruction task, respectively.

\subsubsection{Baselines}
We choose  eight baselines to prove the validity of the common causal knowledge and the advantage of our methods. The baselines are divided into two categories. The first category does not consider the impact of missing features on the prediction of OD flow in data-scarce cities, while the second category reconstructs the missing features first and then performs OD flow prediction.

The following baselines are belong to the first category. These methods use only the observed urban regional features in the data-scarce city with a very small amount of OD flow data to make prediction.
\begin{itemize}
    \item \textbf{Gravity Model \cite{lenormand2016systematic}.} It describes the OD flow between two regions as gravity, where the regions are considered as celestial bodies and the region attributes are considered as its mass. 
    \item \textbf{GBRT \cite{robinson2018machine} .}It combines the gradient boost techniques and decision trees to predict OD flow.
    \item \textbf{GAT \cite{velivckovic2017graph}.} It models the spatial dependencies among regions via GAT (graph attention networks)  and then makes OD flow prediction. 
    \item \textbf{GMEL \cite{liu2020learning}.} It integrates GAT and multi-task learning strategy to extract the geo-contextual embedding of regions and uses GBRT \cite{robinson2018machine} to predict the OD flow between two regions.
\end{itemize}

The second category includes three baselines. They all  reconstruct the missing features first and then make OD flow predictions.
\begin{itemize}
    \item \textbf{MF-GAT.} It tries to reconstruct the missing features case by case through MF (Matrix Factorization) \cite{Koren2009MatrixFT} and then combines the observed features and reconstructed features to predict OD flow via GAT.
    \item \textbf{AE-GAT.} It first trains feature reconstruction model based on an auto encoder (AE) framework~\cite{semwal2017robust,ng2011sparse} in data-rich cities and then  reconstructs the missing features in data-scarce cities. The observed features and reconstructed features are combined to predict OD flow via GAT.
    \item \textbf{VAE-GAT.} It first trains feature reconstruction model based on the variational  auto encoder (VAE)~\cite{lopez2017conditional,sohn2015learning} in data-rich cities and then  reconstructs the missing features in data-scarce cities. The observed features and reconstructed features are combined to predict OD flow via GAT.
\end{itemize}

\subsection{Results Analysis}
In this section, we will present the experimental results of all baselines and our proposed method CE-OFP and give a systematic analysis.
\textbf{Overall OD Flow Prediction Results.}
We summarize the performance of all models in  \textbf{NYC-Chi} and \textbf{NYC-Sea} scenarios as shown in Table \ref{tab:overall}. From Table \ref{tab:overall} we can see that CE-OFP achieves the best performance on all metrics in both two scenarios and reduces
the RMSE of OD flow prediction for data-scarce cities by up to 11\%. What's more,
the methods that reconstruct the missing features in the data-scarce cities have a greater advantage in performance in both scenarios.  In the baselines without feature reconstruction, the gravity model behaves the worst because it fails to well extract the  information of urban regional features. In contrast to this, GMEL achieves the best performance for that it well models the spatial dependence between  urban regions and extracts the geo-contextual embedding of regions for prediction. In the baselines considering feature reconstruction, VAE-GAT shows the best performance because VAE models the accuracy and uncertainty of missing features at the same time.

\textbf{Effect of CE-VAE on OD Prediction under Different Number of Missing Features.}
We  experimentally study the effect of CE-VAE on OD flow prediction under different number of missing features. We draw the changes of RMSE of OD flow prediction with the number of missing features and compare the performance of CE-OFP (ours) and CE-OFP without CE-VAE (missing)  in the \textbf{NYC-Chi} scenario in  Fig.\ref{Fig:missing_num} in Appendix E.  From the blue line in Fig. \ref{Fig:missing_num}  we can see that as the number of missing features gradually increases (from 10 to 50), the prediction performance of the model tends to decrease significantly. As can be seen from the yellow line in Fig. \ref{Fig:missing_num} provided, with the help of CE-VAE learned in the data-scarce city, the performance degradation can be seen to nearly disappear. This further demonstrates the effectiveness and robustness of CE-VAE.


\subsection{Causal Knowledge Analysis}
In this section, we will provide a comprehensive insight into the discovered causal structure among regional feature from the source city, New York City, and the results of approach with respect to causal knowledge modeling.

\subsubsection{Causal Graph Analysis}
\begin{figure}[h]
  \centering
  \includegraphics[width=0.9\linewidth]{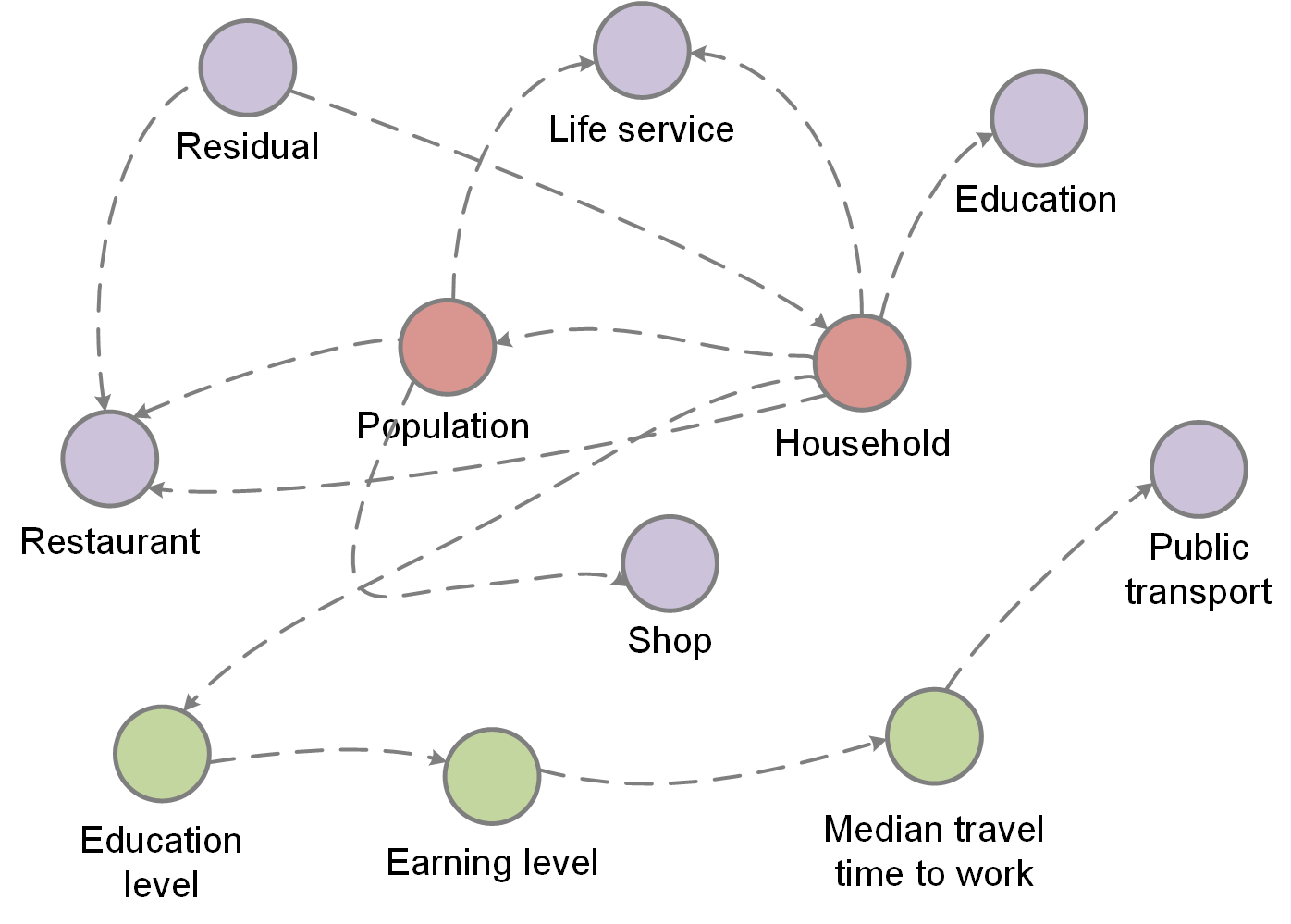}
  \caption{Sub causal graph sampled from the complete causal graph found in New York City. The partial results are convenient for checkinging the reliability of the causal graph from intuition.}
  \label{Fig:sub-causalgraph}
  \vspace{-2mm}
\end{figure}

For experimental purposes, we first verified the generality of the causal structure, which is the basis on which we can transfer between cities. The experimental results show that the causal structure has considerable similarity between different cities, with more than 50\% similarity between them. We provide causal graphs for several cities in the appendix, as shown in Fig. \ref{fig:CDC}.
To facilitate intuitively determining whether the causal graph is reliable, we sampled a subgraph from the causal graph of New York City and displayed it in Fig. \ref{Fig:sub-causalgraph}. From the figure we can see that the causal graph is intuitive. For example, the number of people as the main influence will affect the construction of POI, and the education level will also affect people's income.

\subsubsection{Evaluation of CE-VAE}
In order to verify the reconstruction effect of CE-VAE on missing features, we select log likelihood (L) and MSE metrics to measure the difference between the reconstructed features and the real features. We compare CE-VAE with two other baselines in  feature reconstruction  experiment with 40 missing features in two scenarios, as shown in Table \ref{tab:Evaluation of CEVAE}. It can be seen from the table that CE-VAE shows the best performance for that it models the accuracy and uncertainty of missing features via mean and variance. 

\begin{figure}[h]
  \centering
  \includegraphics[width=0.9\linewidth]{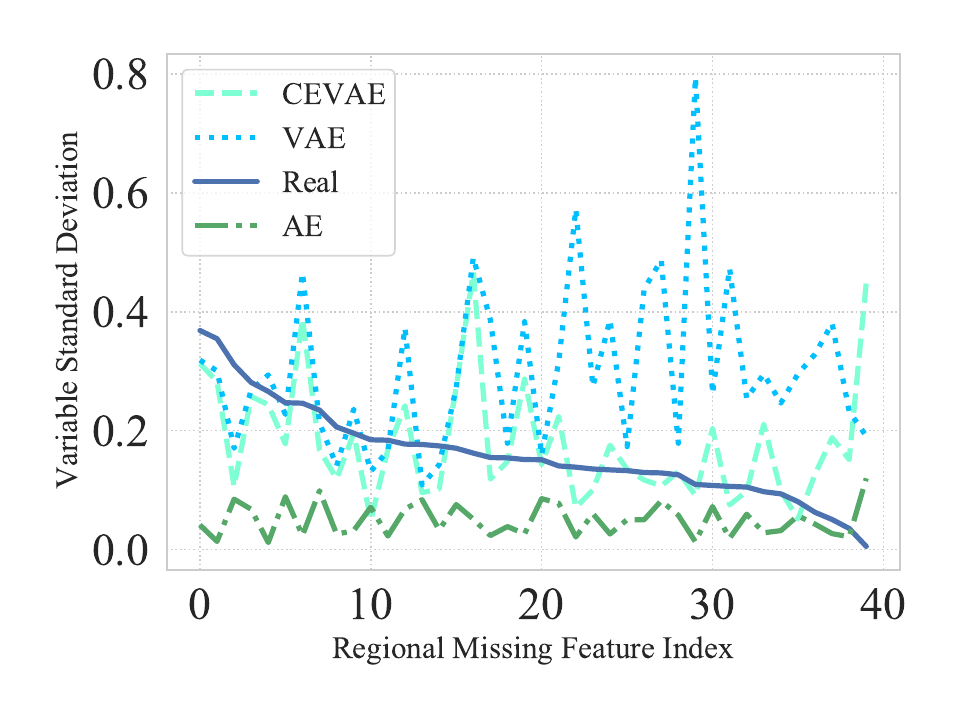}
  \caption{Comparison of the standard deviation between the reconstructed features and the real features. The regional feature index from left to right is sorted according to the standard deviation of the real missing features from large to small.}
  \label{Fig:standard deviation}
  \vspace{-2mm}
\end{figure}

We also draw Fig. \ref{Fig:standard deviation} to compare  the standard deviation of reconstructed features and the real features of our method and baselines. From this figure, it can be found that CE-VAE best reconstructs the standard deviation of real features. From Table \ref{tab:Evaluation of CEVAE} and Fig. \ref{Fig:standard deviation}, we can also find that although AE is more accurate than VAE in its value estimation of real features, its standard deviation is estimated to be worse than VAE, which leads to a worse performance on subsequent prediction tasks as shown in Table \ref{tab:overall}. This is because VAE  models the mean and variance of missing features, thereby obtaining more 
information about  missing features than AE, which only estimates the  mean of missing features.

In addition, we also draw the changes of $L$ and MSE of CE-VAE and VAE during the training process in NYC-Chi (see Fig. \ref{fig:training process in Chi} of Appendix D in detail). Through this figure, we can find that both CE-VAE and VAE will overfit during the training process. However, CE-VAE's verification set can capture this phenomenon in time and curb the further deterioration of the feature reconstruction effect, which makes CE-VAE perform better in subsequent OD flow prediction task. This is because CE-VAE acquires the prior knowledge of the feature relationship of the data-scarce city by constructing the causal  graph between features, so that it can reconstruct the missing features more accurately.

\begin{table}[t]
    \caption{Performance of feature reconstruction.}
    \resizebox{0.3\textwidth}{!}
  {%
    \begin{tabular}{c||cc|cc}
        \Xhline{1pt} 
        \multirow{3}{*}{Model} 
        & \multicolumn{2}{c|}{NYC-Chi}
         & \multicolumn{2}{c}{NYC-Sea}
        \\
         & $L$ $\uparrow$ & MSE $\downarrow$  & $L$ $\uparrow$ & MSE $\downarrow$ \\ 
         \hline \hline
        {AE} &$\times$&0.055&$\times$&0.077  \\
        {VAE} &<-100&0.072&<-100&0.245 \\
        {CE-VAE} &\textbf{14.73}&\textbf{0.054}&\textbf{-20.14}&\textbf{0.074}  \\
         \Xhline{1pt} 
    \end{tabular}}
    \label{tab:Evaluation of CEVAE}
\end{table}

\subsection{Ablation Study}

\begin{table}[]
\centering
\caption{Ablation study in \textbf{NYC-Chi}}
\resizebox{0.35\textwidth}{!}{
\begin{tabular}{@{}llll@{}}
\toprule
Model                          & RMSE $\downarrow$  & SMAPE $\downarrow$ & CPC $\uparrow$ \\ \midrule
CE-OFP (-$CE-VAE$)                       & 10.89     & 0.83      & 0.57    \\
CE-OFP (-$\mathcal{G}$)           & 9.39     & 0.73      &  0.58   \\
CE-OFP (-$\mathcal{L}_{u}$)                  & 9.01     & 0.72      &  0.59   \\
CE-OFP (-$\sigma_z$) &  8.92    & 0.71      & 0.61    \\
CE-OFP & \textbf{8.86}  & \textbf{0.69}   &  \textbf{0.62}   \\ \bottomrule
\end{tabular}}
\label{exp:ablation}
\end{table}

To provide a comprehensive understanding of the key components
of our framework, we conduct a series of experiments to investigate the effect of different components in NYC-Chi scenario. We use the CE-OFP (-$CE-VAE$) as the basic model, which removes CE-VAE from CE-OFP. All methods in this section share the same hyper-parameter settings introduced in Section \ref{sec:hyper-parameter}. The experimental results are summarized in Table \ref{exp:ablation}.

\textbf{CE-OFP (-$\mathcal{G}$).} We first evaluate the effect of feature reconstruction without causal knowledge. This experiment helps us to verify whether it is feasible to use VAE, based on partial observed features to reconstruct the complete features, to learn the representative embedding of urban units. From the result shown in Table \ref{exp:ablation}, the naive reconstruction with vanilla VAE can bring a 8.9\% performance improvement. This is because by modeling the distribution relationship from partial observation to completion, VAE can learn the regional profile representation that contains complete features.


\textbf{CE-OFP (-$\mathcal{L}_{u}$).}  This experiment is employed to check the validity of supervised training of feature estimation in both encoder and decoder via adding auxiliary loss. We explore effect of the estimated expectation of the missing features. From the experimental results, this part of the supervised loss improves the performance. This is because supervised training allows for physically meaningful constraints on the hidden embeddings that model unobserved features. What's more, the physical constraints in the decoder can weaken the error accumulation that comes with the causal pathway.

\textbf{CE-OFP (-$\sigma_z$).}  We investigate the effect of modeling the uncertainty of the missing features via variance, and from the results of this experiment, it is clear that knowing the variance distribution will have more information gain for prediction than only knowing the mean value.

\subsection{Hyper-parameters Study}

\begin{figure}[htbp]
\vspace{-0.2cm}
\centering
\subfigure[Effect of dimention of latent variable $Z$ ]{
    \label{pzdim}
    \includegraphics[width=0.22\textwidth]{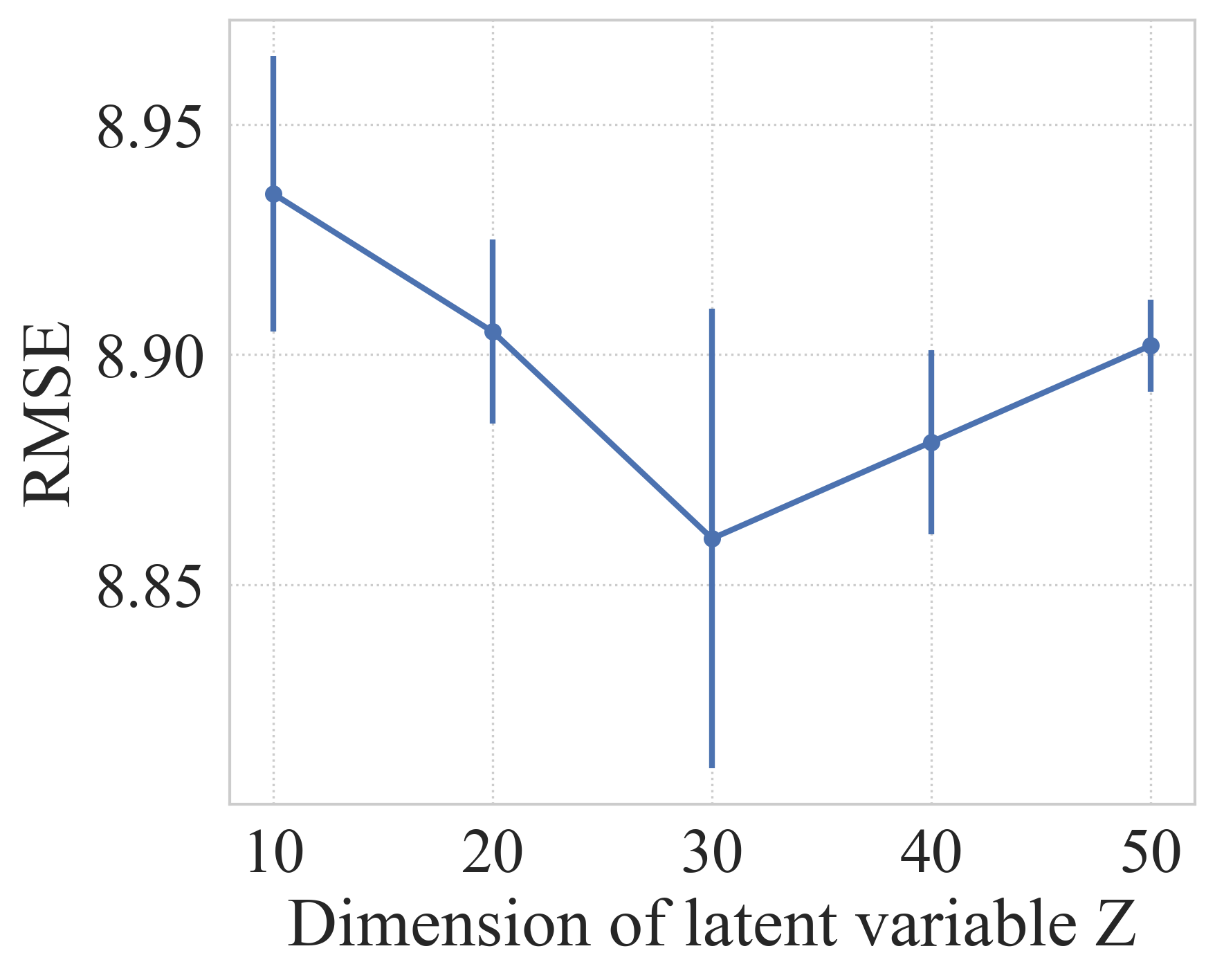}
}
\subfigure[Effect of weight of auxiliary loss $\beta$ ]{
    \label{beta}
    \includegraphics[width=0.22\textwidth]{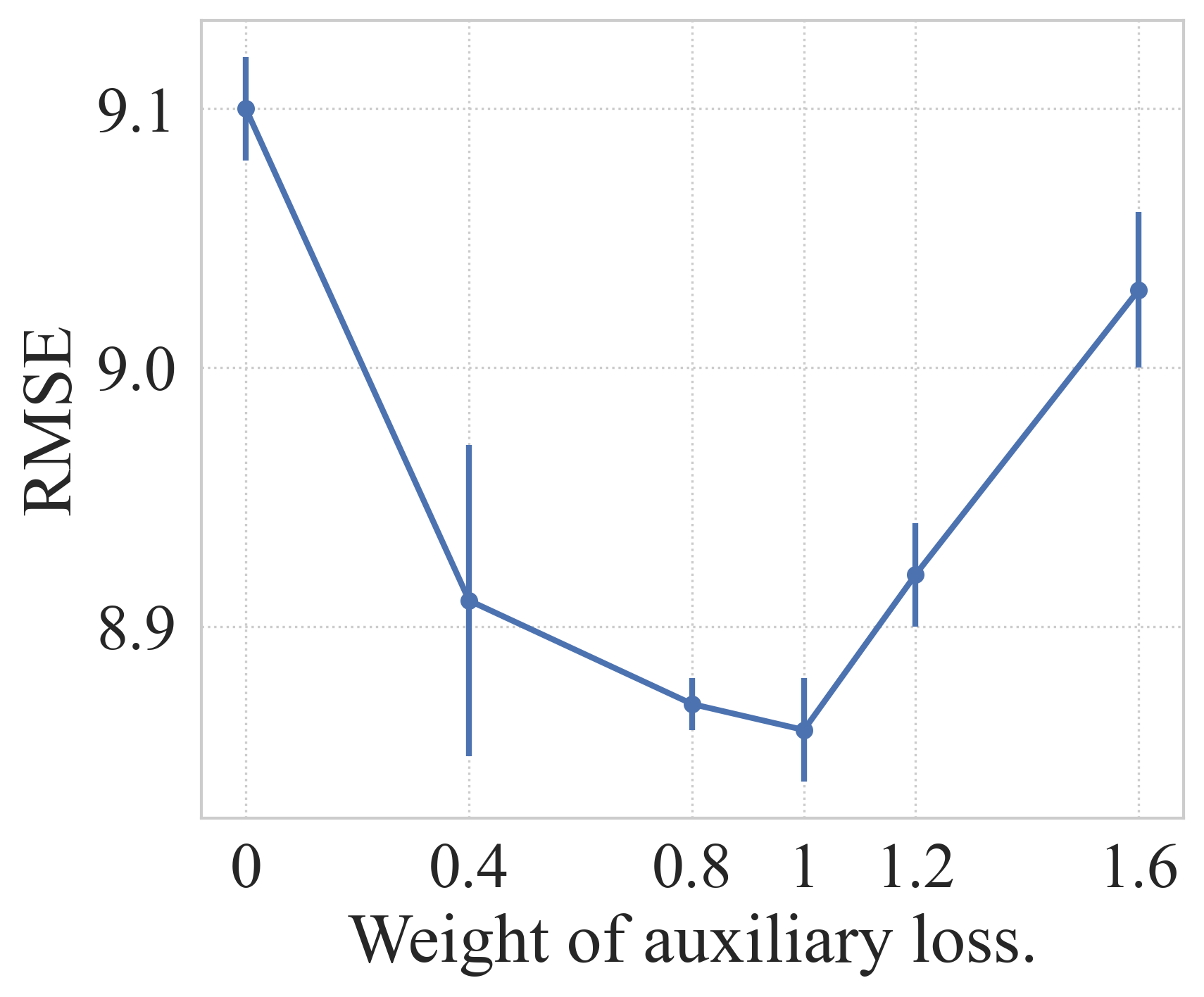}
}

\subfigure[Effect of number of layers in GAT]{
    \label{layers}
    \includegraphics[width=0.22\textwidth]{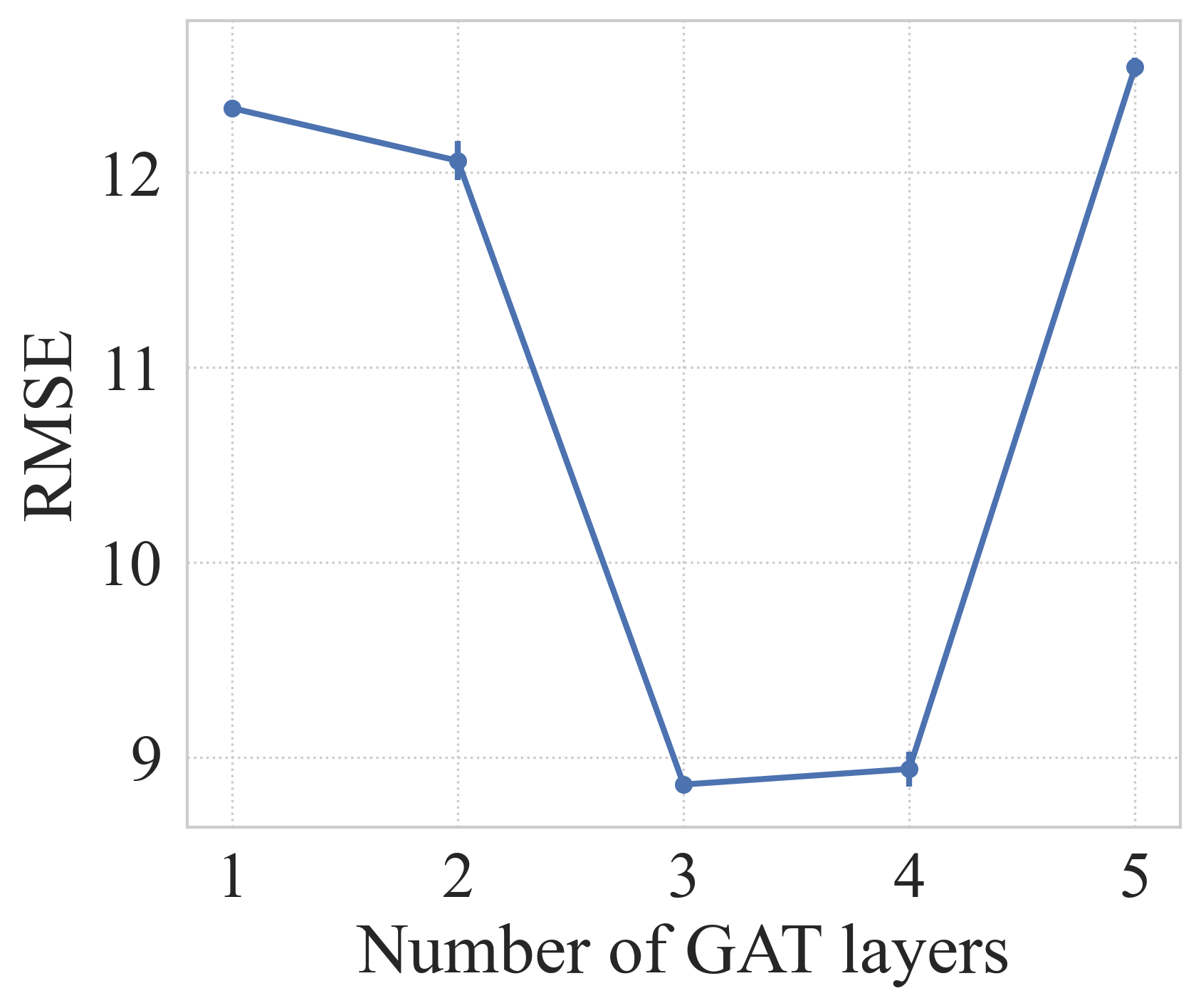}
}
\subfigure[Effect of number of filters of the regional embedding from the GAT]{
    \label{filters}
    \includegraphics[width=0.22\textwidth]{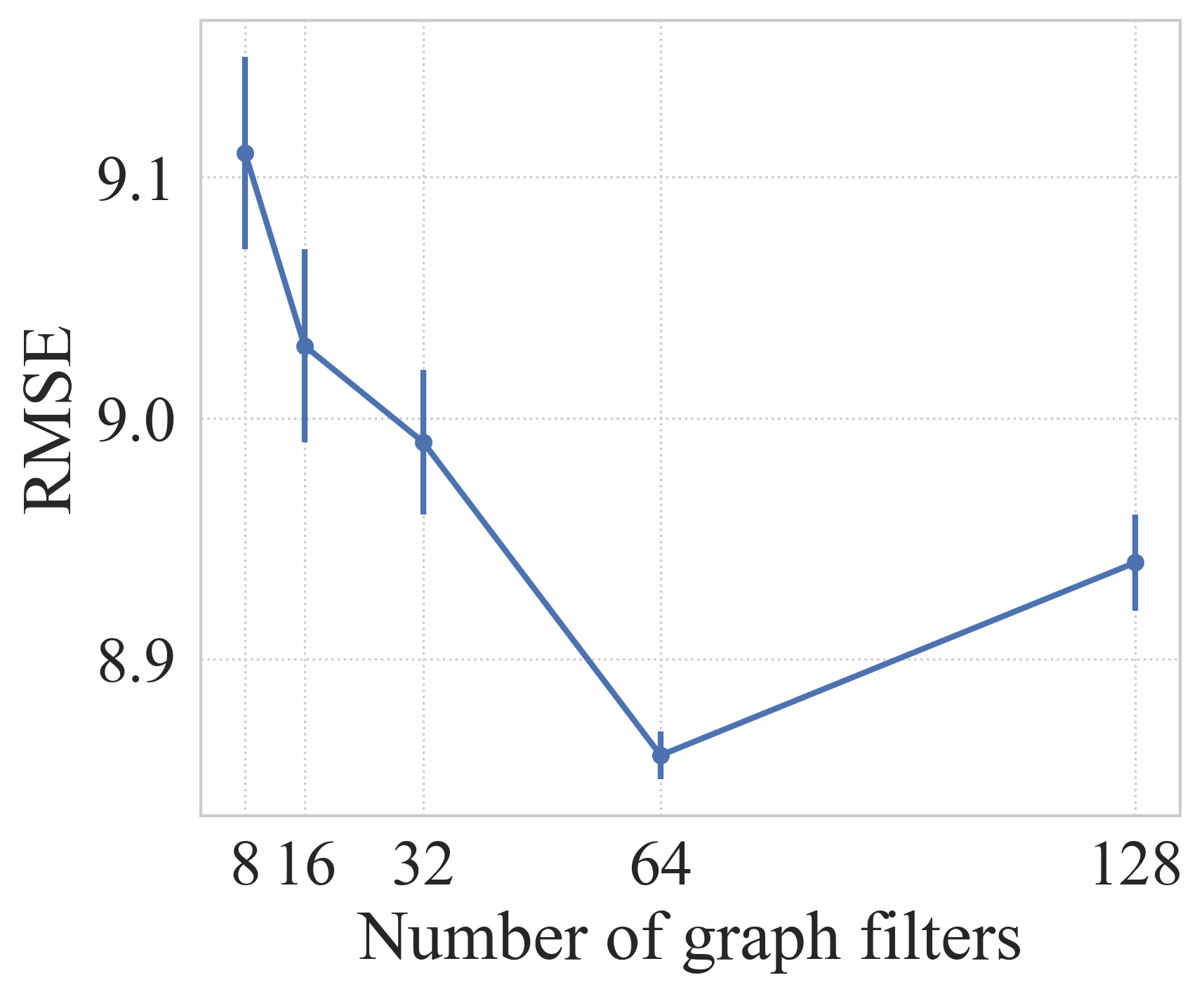}
}
\vspace{-0.2cm}
\caption{Effects of hyper-parameters.}
\vspace{-0.2cm}
\label{ParametersEffect}
\end{figure}

In this section, we conduct experiments to analyze the effect of hyper-parameters. There are four key hyper-parameters: the dimension the latent variable $Z$ of Causal Enhanced Variational Auto-encoder, the weight of auxiliary loss $\beta$, the number of layers in GAT and the number of filters of the regional embedding from the GAT. 
Fig. \ref{pzdim} shows the impact of dimention of latent variable $Z$, where the y-axis is the prediction error, and the x-axis is the dimension of $Z$. Results show that choosing the proper dimension of latent variables $Z$ is important, and both too high and too low dimensions will bring performance degradation.. 
Fig. \ref{beta} shows the impact of the weight of auxiliary loss $\beta$, where the y-axis is the prediction error, and the x-axis is the weight size of $\beta$. Results show that ELBo loss and auxiliary loss are both very important, and a suitable trade off between them is needed to achieve the best performance.
Fig. \ref{layers} shows the impact of the number of layers in GAT, where the y-axis is the prediction error, and the x-axis is the layer number. Results show that the number of layers of the graph neural network has a significant impact on the prediction performance. Shallow layers cannot propagate the information effectively, but deeper layers will have the over-smoothing phenomenon of graph neural networks, which brings performance loss.. 
Fig. \ref{filters} shows the impact of the number of filters of the regional embedding from the GAT, where the y-axis is the prediction error, and the x-axis is the number of filters. Results show that number of filters will have a great effect of the prediction performance. 

In a nutshell, both the modeling of urban causal knowledge and prediction model require consideration of the number of parameters. A model with  a small number of parameters cannot model the complex causal relationships between numerous urban factors, while a large number of parameters can easily lead to overfiting and reduce the generalization ability of the model.

\section{Related Works} \label{Sec:related}


\para{OD Flow Prediction.} 
Research on OD flow prediction. has a very long history, and recent research is very active. Classical works \cite{carey1871principles,stouffer1940intervening,simini2012universal} tend to mimic the physical laws to model population movement in the city. Traditional methods \cite{carey1871principles,stouffer1940intervening,simini2012universal} model the flow of people between regions via simple physical processes, but they cannot perform well in practice due to their poor expressiveness and limited accuracy. With the rapid development of machine learning and deep learning, decision trees \cite{safavian1991survey} and graph neural networks \cite{velivckovic2017graph} show an advantage in predicting OD flow. Nonetheless, these approaches re-
quire a large amount of OD flow data to fit the large number of
model parameters, which prevents their applications in data-scarce
cities. The recent transfer learning techniques \cite{robinson2018machine,pourebrahim2019trip} provide promising paths to OD flow prediction by transferring data and model from data-rich city to data-scarce  city, but the missing regional features  are the bottleneck of them.
Based on the above research, we propose a novel Causality-Enhanced OD Flow Prediction (CE-OFP), a unified framework that aims to transfer urban knowledge between cities and achieve accuracy improvements in OD flow predictions across data-scarce cities.

\para{Causal Knowledge Modeling.}
The core of causal knowledge modeling is to search the causal graph through causal discovery methods. There are many researches on causal discovery. Traditional methods are often based on Bayesian inference~\cite{solus2017consistency,raskutti2018learning,schmidt2007learning,teyssier2012ordering} or linear programming~\cite{bartlett2017integer,xiang2013ast}.    Although these methods can search  the causal graph, they need to be improved in accuracy and efficiency.  Recently, RL-based methods have achieved promising result in causal discovery with large scale nodes, which regard causal discovery  as a combinatorial optimization problem. Zhu et al.~\cite{zhu2019causal} model the causal discovery as graph generation problem and utilize RL to help search the best directed acyclic graph (DAG) by minimizing the Bayesian Information Criterion (BIC)~\cite{watanabe2013widely,wang2021ordering}. 
In our work, we regard  urban regional features as nodes of causal graph and utilize the state-of-art RL-based method ~\cite{watanabe2013widely} to help search the causal relations between regional features, therefore extracting the urban knowledge for downstream tasks including feature reconstruction and OD flow prediction.


\para{Feature Reconstruction.}
Feature reconstruction plays an important role in prediction tasks. Some studies have shown that feature reconstruction can help improve the effect of prediction. Isogawa et al.~\cite{isogawa2017image,jiang2019robust} exploit time series data to complement missing information on people's moving images and use them in the task of human movement prediction based on the framework of LSTM, thereby enhancing the prediction accuracy of people's movement. Semwal et al.~\cite{semwal2017robust} utilizes known physical characteristics to reconstruct humanoid push based on auto encoder (AE)~\cite{ng2011sparse} and variational auto encoder (VAE)~\cite{kingma2013auto,sohn2015learning,lopez2017conditional}, which improves the accuracy of predicting physical movements. Inspired by the above research, based on the framework of VAE, we obtain the relationship path between the collected features and the missing features through the obtained urban causal graph, and train the feature reconstruction model in the data-rich cities and migrate it to the developing cities with scarce features.


\section{Conclusion}~\label{sec:con}
In this study,  we have explored to leverage generalized urban knowledge from data-rich cities
to compensate data scarce issues in developing cities for accurate OD flow prediction.
Different from previous tasks of performing OD flow prediction only for single-city scenarios,
our key contribution is to propose a unified framework CE-OFP that aims to transfer urban knowledge
between cities and achieve accuracy improvements in OD flow predictions across data-scarce cities. Experimental results have shown that our proposed CE-OFP framework can significantly improves prediction accuracy of OD flows in data-scarce cities, and it outperform seven state-of-the-art baseline methods by up to 11\% in RMSE.

\clearpage
\bibliographystyle{ACM-Reference-Format}
\bibliography{reference}
\clearpage
\appendix
\section{The pseudo algorithm of CE-OFP}

\begin{algorithm}[h]
\caption{Training of CE-OFP}
\label{alg:cevae}
\begin{algorithmic}[1]

\REQUIRE ~~\\
Complete features of all regions in the data-rich  city $\{F_i^r|i=1,2,...,|\mathcal{F}| ~ and ~ r\in \mathcal{R}^{src}  \}$. \\
Complete origin-destination flow matrix of the data-rich city $\mathcal{M}^{src}$. \\
Partially observed features of all regions in the data-scarce cities $\{ X_i^r | i=1,2,...,|\mathcal{X}| ~ and ~ r \in \mathcal{R}^{tar} \}$.\\

\ENSURE ~~\\
Learned CE-VAE. \\
Estimated complete OD flow matrix of the data-scarce cities $\mathcal{M}^{tar}$.\\

//construct the training and validation set based on data from the data-rich city

\STATE sample 90\% of $r \in \mathcal{R}^{src} $ to construct the training set $\mathbb{D}_{training}$
\STATE use 10\% of $r$ left to construct the validation set $\mathbb{D}_{validation}$

\STATE $\mathbb{D}_{training} \Leftarrow \emptyset$

\FOR{all $r \in \mathbb{D}_{training}$}
\STATE $D_i \Leftarrow {\;}   [\textbf{X}^r_{ \{ i=1,2,...,|\mathcal{X}|  \} }, \textbf{F}^r_{ \{ i=1,2,..,|\mathcal{F}| \} } ] $
\STATE put $D_i$ into ${\mathbb{D}_{training}}$
\ENDFOR  \\
//train CE-VAE model
\STATE initialize the learnable parameters $\theta$ of CE-VAE
\REPEAT
    \STATE choose a batch of $D$ from $\mathbb{D}$
    \STATE compute the loss function $\mathcal{O}$ using (11)
    \STATE optimize $\theta$ by Adam to minimize the loss function $\mathcal{O}$ based on the choosen batch of $D$
    \STATE compute the $\mathcal{L}_{u}$ of $\mathbb{D}+{validation}$
\UNTIL{loss $\mathcal{O}$ converge} \\

//train GAT-based OD Flow Prediction Model via
Knowledge Distillation 

\STATE Obtain $\mu_z$ and $\sigma_z$ from the encoder of  learned CE-VAE and combine them with the original regional features to construct the  training and validation set in data-scarce city as $\mathbb{D'}_{training}$  and $\mathbb{D'}+{validation}$

\STATE initialize the learnable parameters $s$ and $t$ for GAT models of data-rich city and data-scarce city
\REPEAT
    \STATE choose a batch of $D'$ from $\mathbb{D'}$
    \STATE compute the loss function $\mathcal{L}_{pred}$ using (17)
    \STATE optimize $s$  and  $t$ by Adam to minimize the loss function $\mathcal{L}_{pred}$ based on the choosen batch of $D'$
\UNTIL{loss $\mathcal{L}_{pred}$ converge} \\

\end{algorithmic}
\label{tab:all}
\end{algorithm}

\section{Parameter Settings} \label{sec:hyper-parameter}
We discuss the hyper-parameters setting in this section. We set up a scenario with 40 missing features and 10000 origin-destination flows to perform a performance comparison experiment with all baselines. There isn't any hyper-parameters in gravity model. The number of estimators of GBRT is set to 100 with no increment on performance by increasing this hyper-parameter. The GNN (graph neural networks) based models, including GAT, GMEL, MF-GAT, AE-GAT, VAE-GAT and our proposed method, all have 3 layers and 64 filters. The training termination condition for all models is either loss convergence or overfitting on the validation set.

\section{Causal Knowledge Modeling}



\subsection{Training Process of Causal Discovery Algorithm}
We exploit a RL-based causal discovery algorithm and we plot its training process in  Fig. 11. The figure depicts the change of reward with the number of training steps. It can be found that the reward gradually increases with the training until it converges.

\subsection{Discovered Causal Graph}
We exploit a RL-based causal discovery algorithm to discover the causal graph in four cities as shown Fig.8. It can be found that these causal graphs have many similarities and overlaps, which demonstrates the generalization of causal knowledge in cross-city scenarios.

\begin{figure}[h]
\subfigure[Causal graph of NYC.]{\includegraphics[height=0.19\textwidth]{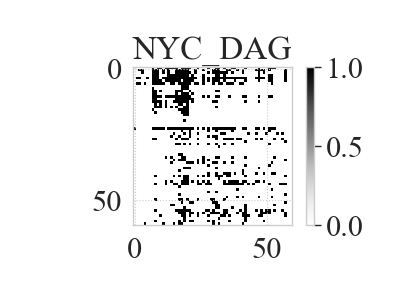}}\subfigure[Causal graph of Chi. ]{\includegraphics[height=0.19\textwidth]{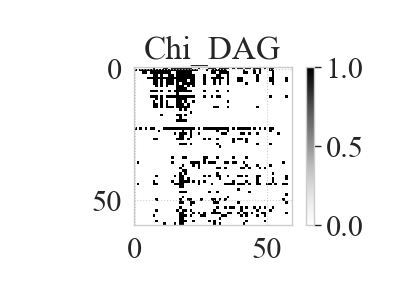}}
\subfigure[Causal graph of LA.]{\includegraphics[height=0.19\textwidth]{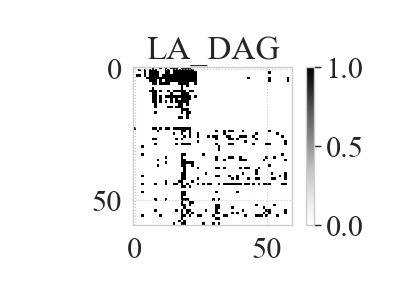}}\subfigure[Causal graph of Sea. ]{\includegraphics[height=0.19\textwidth]{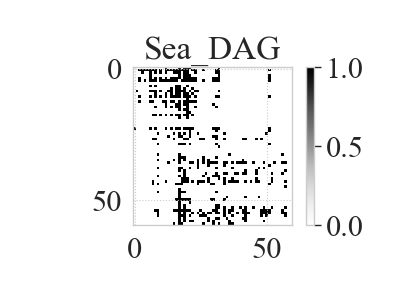}}
\caption{Discovered causal graph of four cities.}
\label{fig:CDC}
\end{figure}

\section{Training Process Comparison of CE-VAE and VAE}

We plot the training process of CE-VAE and VAE in  Fig. 9. Specifically, we plot the MSE and log-likelihood trends of the two models in NYC (data-rich city) as the validation set and Chi (data-scarce city) as the test set. From the error correspondence between the verification set and the test set in the figure, it can be found that the CE-VAE method can detect the overfitting trend of the test set in time on the verification set, thereby preventing overfitting. However, VAE is easy to overfit; therefore, VAE achieves worse results than CE-VAE in cross-city feature reconstruction. The above findings also verify the generalization ability of CE-VAE.

\begin{figure}[]
\subfigure[The changes of MSE in CE-VAE training process in NYC-Chi.]{\includegraphics[height=0.17\textwidth]{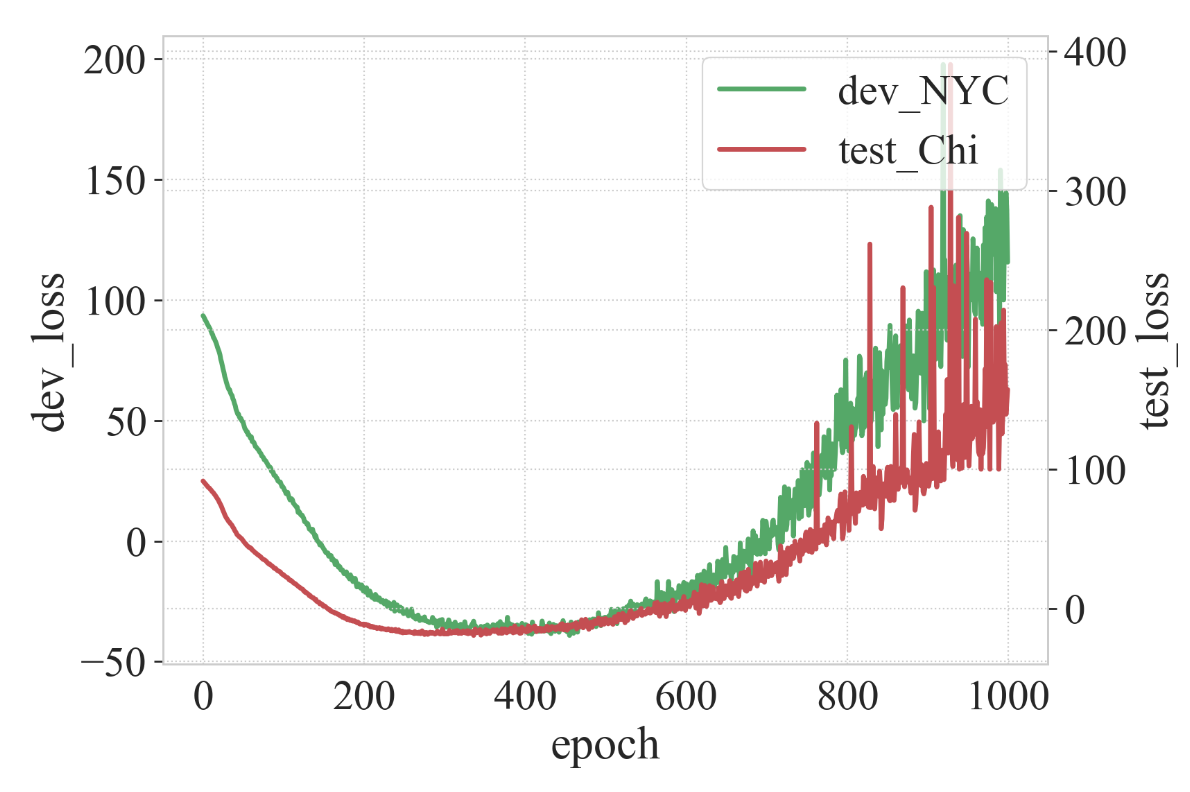}}\subfigure[The changes of log-likelihood in CE-VAE training process in NYC-Chi. ]{\includegraphics[height=0.17\textwidth]{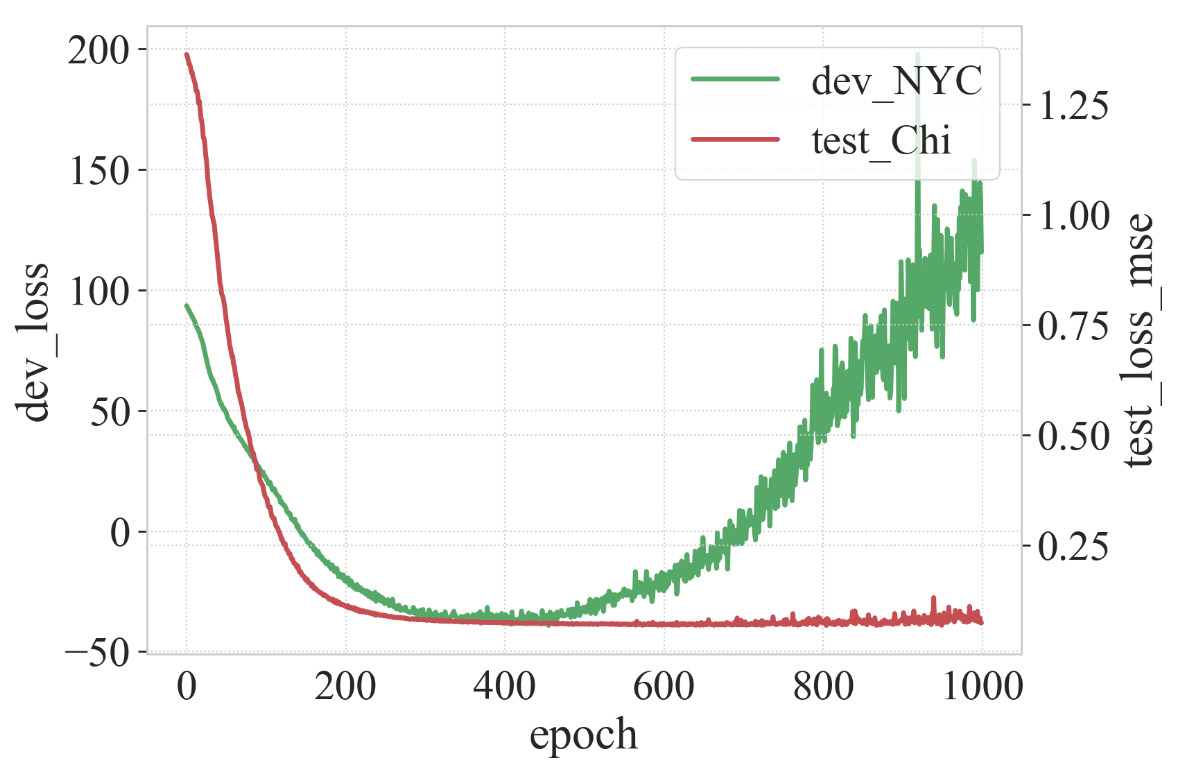}}
\subfigure[The changes of MSE in VAE training process in NYC-Chi.]{\includegraphics[height=0.17\textwidth]{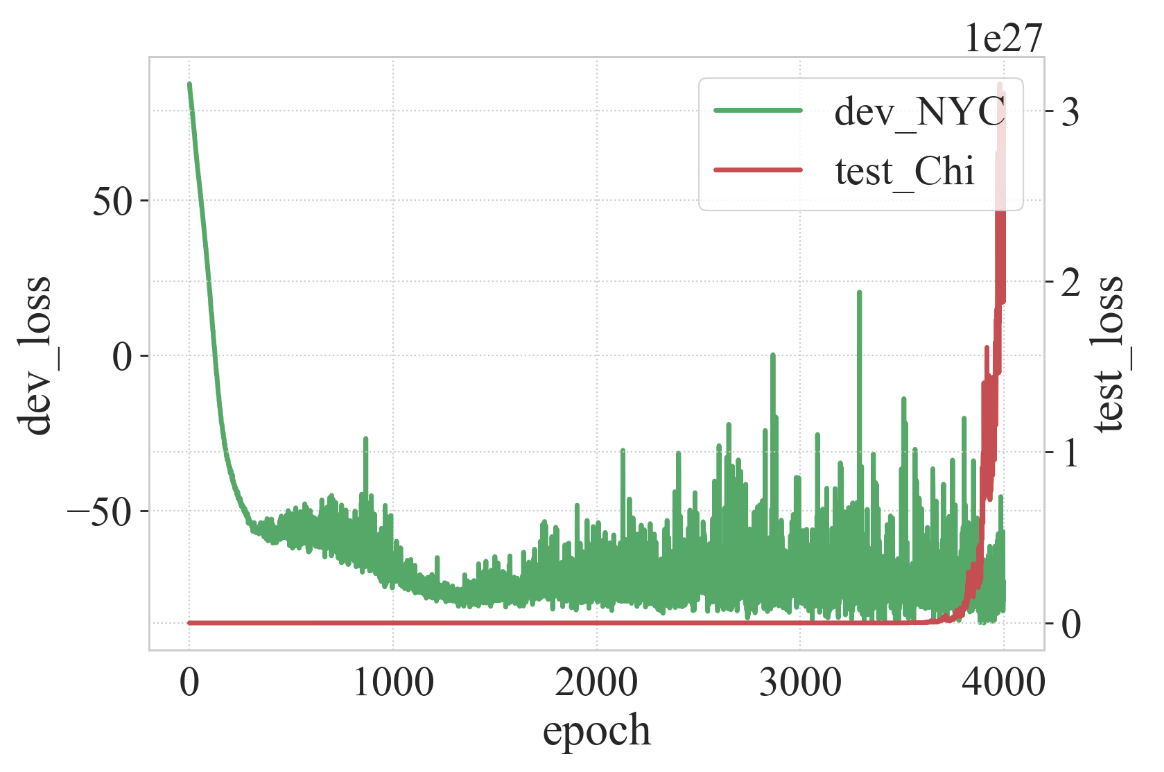}}\subfigure[The changes of log-likelihood in VAE training process in NYC-Chi. ]{\includegraphics[height=0.17\textwidth]{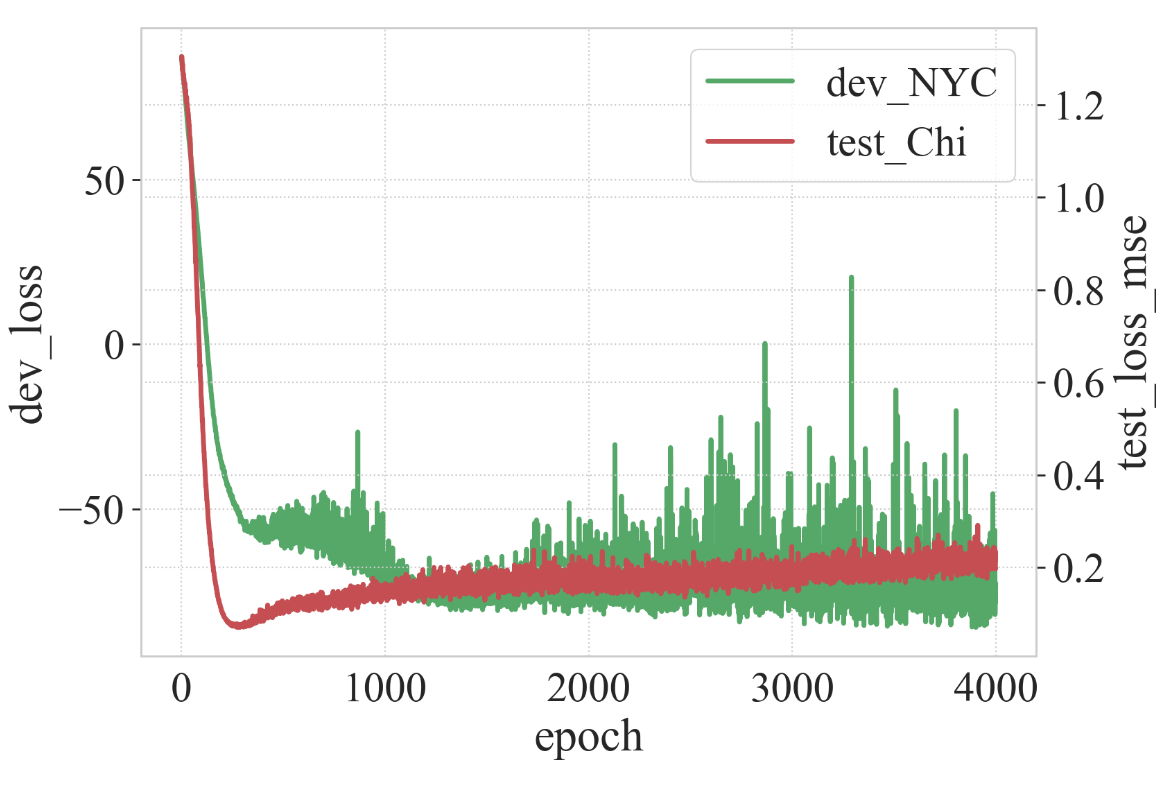}}
\caption{Loss changes  of the CEVAE-based and VAE-based feature recovery model in the NYC's validation set and Chi's  test set.}
\label{fig:training process in Chi}
\end{figure}

\section{Effect of CE-OFP under different number of missing features}

We plot prediction RMSE of CE-OFP and CE-OFP(-CE-VAE) under different number of missing features in NYC-Chi as shown in Fig. 10. It can be seen from the figure that the more features missing in Chi, the worse the prediction performance of CE-OFP (-CE-VAE) is, while the prediction performance of CE-OFP remains good. This demonstrates the effectiveness and robustness of CE-OFP.
\begin{figure}[]
  \centering
  \includegraphics[width=0.7\linewidth]{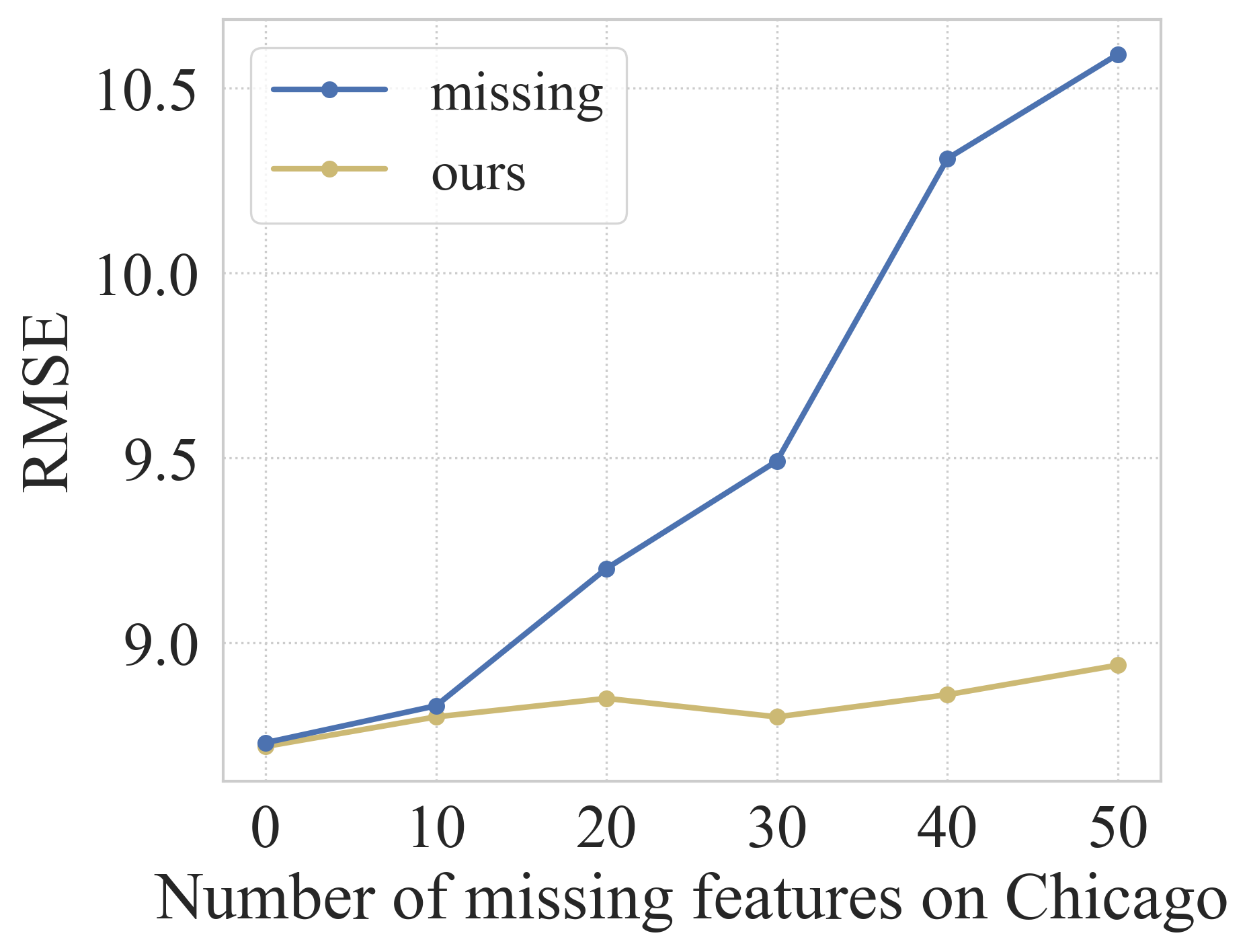}
  \caption{Prediction RMSE of CE-OFP under different number of missing features in NYC-Chi.}
  \label{Fig:missing_num}
\end{figure}

\begin{figure}[]
  \centering
  \includegraphics[width=0.7\linewidth]{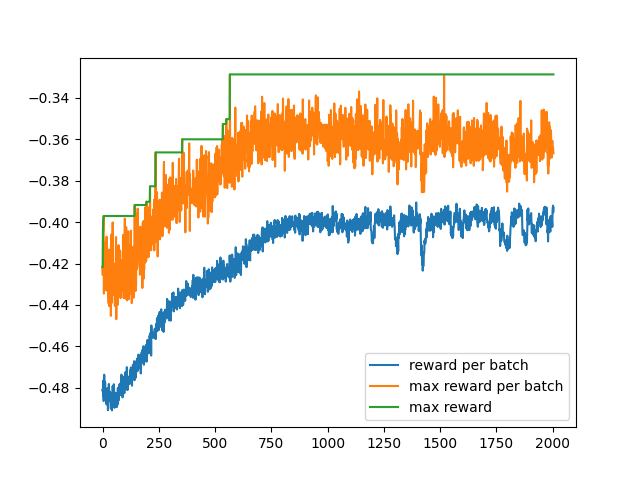}
  \caption{The changes of reward during the training process of RL-based causal discovery.}
  \label{Fig:cvae_training}
\end{figure}


\end{document}